\documentclass[letterpaper]{article} 
\usepackage{aaai2026}  
\usepackage{times}  
\usepackage{helvet}  
\usepackage{courier}  
\usepackage[hyphens]{url}  
\usepackage{graphicx} 
\usepackage{natbib}  
\usepackage{caption} 
\usepackage{algorithm}
\usepackage{algorithmic}

\usepackage{newfloat}
\usepackage{listings}
\DeclareCaptionStyle{ruled}{labelfont=normalfont,labelsep=colon,strut=off} 
\floatstyle{ruled}
\newfloat{listing}{tb}{lst}{}
\floatname{listing}{Listing}
\usepackage[utf8]{inputenc} 
\usepackage{booktabs}       
\usepackage{amsfonts}       
\usepackage{nicefrac}       
\usepackage{microtype}      
\usepackage{xcolor}         
\usepackage[table]{xcolor}
\usepackage[most]{tcolorbox}

\usepackage{graphicx}
\usepackage{xcolor}
\usepackage{amssymb}
\usepackage{amsmath}
\usepackage{soul}

\usepackage{amsthm}
\newtheorem{proposition}{Proposition}

\newcommand{\xmark}{\textcolor{red}{$\times$}}
\newcommand{\cmark}{\textcolor{blue}{$\checkmark$}}

\newtcolorbox{myBox}[1][]{
    enhanced,
    title=#1,
    breakable = true,
    overlay={%
        \ifcase\tcbsegmentstate
        \or%
        \else%
        \fi%
    }
    }

\title{Rethinking Progression of Memory State in Robotic Manipulation:\\ An Object-Centric Perspective}
\author{
    Nhat Chung\textsuperscript{\rm 1}\equalcontrib,
    Taisei Hanyu\textsuperscript{\rm 2}\equalcontrib,
    Toan Nguyen\textsuperscript{\rm 1},
    Huy Le\textsuperscript{\rm 1},\\
    Frederick Bumgarner\textsuperscript{\rm 2},
    Duy Minh Ho Nguyen\textsuperscript{\rm 3,7,8},
    Khoa Vo\textsuperscript{\rm 2},\\
    Kashu Yamazaki\textsuperscript{\rm 4},
    Chase Rainwater\textsuperscript{\rm 2},
    Tung Kieu\textsuperscript{\rm 5},
    Anh Nguyen\textsuperscript{\rm 6},
    Ngan Le\textsuperscript{\rm 2},
}
\affiliations{
    \textsuperscript{\rm 1}FPT Software AI Center\quad
    \textsuperscript{\rm 2}University of Arkansas\\
    \textsuperscript{\rm 3}University of Stuttgart\quad
    \textsuperscript{\rm 4}Carnegie Mellon University\quad
    \textsuperscript{\rm 5}Aalborg University\\
    \textsuperscript{\rm 6}University of Liverpool\quad
    \textsuperscript{\rm 7}German Research Center for Artificial Intelligence (DFKI)\\
    \textsuperscript{\rm 8}Max Planck Research School for Intelligent Systems (IMPRS-IS)\\

    nhatchung14@gmail.com, \{thanyu, thile\}@uark.edu
}

\usepackage{bibentry}

\begin{document}

\maketitle

\begin{abstract}
As embodied agents operate in increasingly complex environments, the ability to perceive, track, and reason about individual object instances over time becomes essential, especially in tasks requiring sequenced interactions with visually similar objects.
In these non-Markovian settings, key decision cues are often hidden in object-specific histories rather than the current scene. Without persistent memory of prior interactions (what has been interacted with, where it has been, or how it has changed) visuomotor policies may fail, repeat past actions, or overlook completed ones.
To surface this challenge, we introduce LIBERO-Mem, a non-Markovian task suite for stress-testing robotic manipulation under object-level partial observability. It combines short- and long-horizon object tracking with temporally sequenced subgoals, requiring reasoning beyond the current frame.
However, vision-language-action (VLA) models often struggle in such settings, with token scaling quickly becoming intractable even for tasks spanning just a few hundred frames.
We propose Embodied-SlotSSM, a slot-centric VLA framework built for temporal scalability. It maintains spatio-temporally consistent slot identities and leverages them through two mechanisms: (1) slot-state-space modeling for reconstructing short-term history, and (2) a relational encoder to align the input tokens with action decoding. 
Together, these components enable {temporally grounded, context-aware action prediction}. Experiments show Embodied-SlotSSM's baseline performance on LIBERO-Mem and general tasks, offering a scalable solution for non-Markovian reasoning in object-centric robotic policies.
\end{abstract}

\begin{links}
    \link{Extended Version}{https://libero-mem.github.io}
\end{links}

\section{Introduction}

\begin{figure*}[t!]
\vspace{0.1cm}
\centerline{\includegraphics[width=0.95\linewidth]{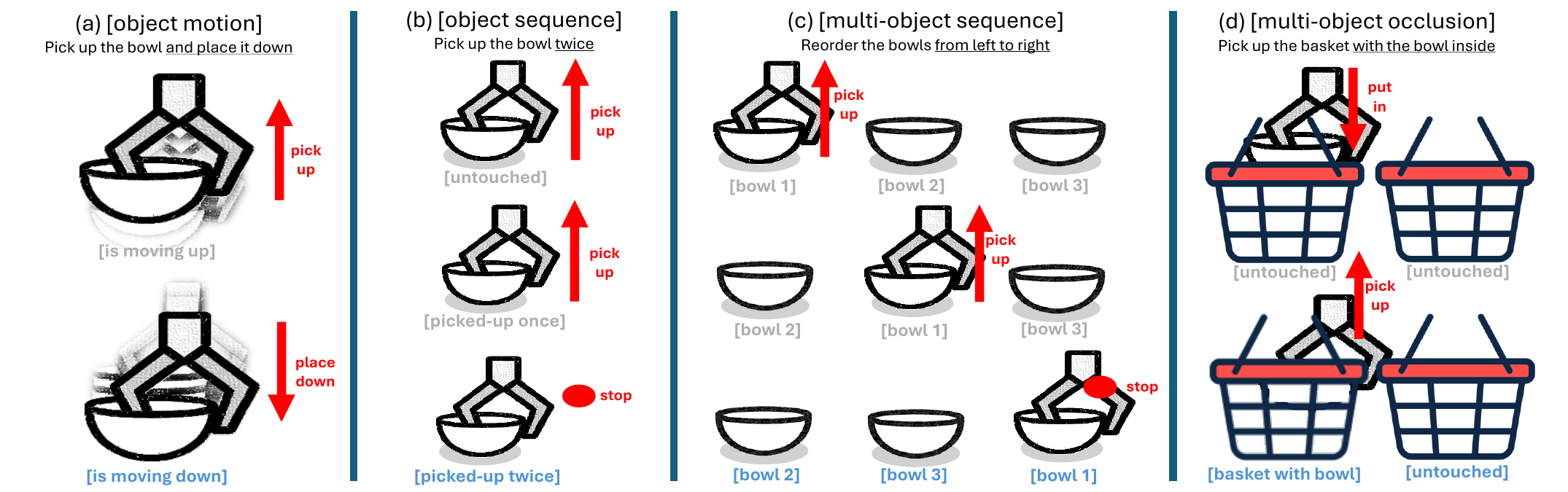}}
    \caption{\textbf{LIBERO-Mem}: robotic manipulation tasks of object-level POMDP dependencies. These tasks require memory of prior actions and object-specific state tracking beyond what purely Markovian or fully observable policies can handle, highlighting the importance of persistent, object-specific memory for short- and long-horizon reasoning across visually similar inputs. In (a) object motion (OM), the robot must recall its last action (e.g., pick up or place down) to act correctly. In (b) object sequence (OS), success depends on remembering how many times an object has been manipulated, since visual cues are insufficient. In (c) multi-object sequence (OR), the robot must track the temporal order of object relations and interactions (e.g., from left to right). In (d) multi-object occlusion (OO), occluded objects require the robot to rely on memory of past placements to identify targets.} 
\label{fig:non-markov-mem}
\end{figure*}

Humans effortlessly recall past interactions with specific objects, such as where they last placed a salt bottle or whether they have already sprinkled salt into a pot of soup a number of times, enabling them to carry out long-horizon, multiple-step tasks with precision. This object-level memory plays a vital role in avoiding redundant actions or missing steps, especially in tasks involving repetitive steps, visually similar items, and extended temporal dependencies. Such object-level challenges are inherently \emph{partially observable Markov decision processes (POMDP)} (illustrated in Fig.~\ref{fig:non-markov-mem}), where the agent's current observation does not fully capture the true state of the environment, and optimal decisions depend on the history of interactions associated with specific objects.

Conversely, robotic visuomotor policies typically rely solely on the most recent sensory inputs to determine subsequent actions ~\cite{DBLP:OpenVLA2024, DBLP:RoboFlamingo2024, DBLP:RoboAgent_2024, DBLP:RT12023, DBLP:Octo2024, zawalski2024robotic, wang2024hpt, DBLP:conf/rss/YangGBSVFSL24, tian2024tokenize}, lacking mechanisms to encode and recall object-centric history \cite{BridgeDatav2_2023, DBLP:RTX2024, DROID2024, james2020rlbench, libero2023, robocasa2024}.
Likewise, most robotic benchmarks~\cite{james2020rlbench, libero2023, robocasa2024, mu2021maniskill, gu2023maniskill2, DBLP:journals/corr/maniskill3_2024} primarily focus on evaluating policies in short-horizon or atomic tasks. Although several benchmarks have emerged to evaluate long-horizon tasks that require memory modeling~\cite{memoryBench2025, MIKASArobo2025, morad2023popgym, chevalier2023minigrid, pleines2025memory}, they largely overlook the challenges posed by POMDP settings.

To address this gap in benchmarking, we introduce \textbf{LIBERO-Mem}, a new suite of manipulation tasks specifically designed to evaluate a model's ability to retain object-centric interactions over time. Unlike prior memory benchmarks, LIBERO-Mem emphasizes object-level memory under ambiguity in object identity, location, and relational history, explicitly targeting complex POMDP scenarios. The suite includes four task types: (a) object motion (OM), (b) object sequence (OS), (c) multi-object sequence (OR), and (d) multi-object occlusion (OO). Each task type is designed to evaluate different aspects of transient and persistent memory by introducing ambiguities that can only be resolved through temporal reasoning over the history of object interactions. By focusing on memory under uncertainty, LIBERO-Mem takes a key step toward enabling general-purpose robots to operate reliably in everyday settings, where even seemingly simple tasks, such as identifying the correct object to manipulate, become challenging without awareness of prior interactions.

As a step towards overcoming the limitations of existing visuomotor policies, we propose \textbf{Embodied-SlotSSM}, a novel VLA framework that integrates slot-based state-space modeling (SSM)~\cite{jiang2024slot} to maintain structured, object-centric memory over time. Embodied-SlotSSM encodes temporal dynamics and inter-object interactions into discrete, persistent slots, enabling consistent tracking of entities throughout a task. This structured memory supports robust state estimation and informed decision-making, particularly in partially observable and non-Markovian settings where access to interaction history is critical. By grounding both perception and action in visual semantics and high-level task intent, Embodied-SlotSSM enables more adaptive and reliable behavior under real-world uncertainty.

To highlight the limitations of existing works and the challenges posed by our benchmark, we evaluate Embodied-SlotSSM and prior state-of-the-art VLA models on the benchmark of the LIBERO-Goal~\cite{libero2023}, and our proposed {LIBERO-Mem} benchmark for POMDP tasks. We also carry out studies of our proposed method in capturing long-term dependencies, improving action prediction, and enhancing manipulation efficiency. Our results highlight the practicality and scalability of slot-based memory representations, paving the way for more reliable and memory-efficient robotic systems in non-Markovian environments.

In summary, our contributions are threefold:

\begin{itemize}
\item We introduce {LIBERO-Mem}, a novel {non-Markovian robotic manipulation benchmark} that systematically evaluates memory-augmented models on long-horizon tasks, emphasizing object permanence, historical reasoning, and structured memory retention.

\item We present {Embodied-SlotSSM}, a {slot-based state-space modeling framework} that encodes persistent, object-centric memory representations, enabling structured tracking and decision-making under partial observability.

\item We conduct {experiments in both general Markovian and special non-Markovian settings} via an oracle-supported implementation of Embodied-SlotSSM, denoted Naive E-SlotSSM, showing that Embodied-SlotSSM enhances stateful reasoning, long-horizon action prediction, and performance on manipulation tasks.

\end{itemize}

We hope this work provides valuable insights and helps bridge the gap between existing VLA designs and the development of memory-centric capabilities in robotics.


\section{Related Works}
\label{sec:related}
\textbf{Robotic Manipulation Benchmarks.} Robotic manipulation has been widely benchmarked across several dimensions, including general task completion~\cite{gu2023maniskill2, kumar2023robohive, fang2023rh20t}, multimodal support~\cite{jiang2023vima, robocasa2024, libero2023}, and sim-to-real generalization~\cite{domainrandom2017, li2024evaluating}. However, many of these benchmarks, such as RLBench~\cite{james2020rlbench}, LIBERO~\cite{libero2023}, and RoboCasa~\cite{robocasa2024}, are constructed under the Markovian assumption, where the robot's next action can be predicted solely from the current observation, without requiring access to historical context.
To address this limitation, recent efforts such as MemoryBench~\cite{memoryBench2025} and MIKASA-Robo~\cite{MIKASArobo2025} have highlighted the importance of memory in robotic manipulation tasks. MemoryBench introduces a limited set of tasks focusing primarily on spatial memory within short-horizon scenarios. MIKASA-Robo expands this scope by incorporating a broader suite of memory types, including object, spatial, and sequential memory across 32 tasks. However, both benchmarks primarily operate under simplified settings and lack object-level ambiguities or temporal scaling. These efforts demonstrate the growing attention toward non-Markovian reasoning, but fall short in systematically stress-testing object-centric memory under compositional and temporally challenging conditions.

In contrast, we propose \textbf{LIBERO-Mem}, a new benchmark explicitly designed to evaluate long-horizon, object-level memory in partially observable robotic settings. LIBERO-Mem features simple yet composable manipulation tasks that require agents to reason over object identity, spatial configuration, and interaction history. Scenarios such as occluded object recall and sequence-dependent pick-and-place (see Fig. \ref{fig:non-markov-mem}) make memory indispensable for success, going beyond what prior benchmarks test. Furthermore, LIBERO-Mem uniquely incorporates stress-testing via temporal scaling, enabling fine-grained assessment of policy robustness over extended episodes - addressing a key gap in existing benchmarks. An overview comparison between our LIBERO-Mem with other benchmarks are summarized in Table 1.

\textbf{VLA Models in Non-Markovian Settings.}
End-to-end VLA models have shown strong performance across various embodied tasks, including scene understanding~\cite{jatavallabhula2023conceptfusion, gu2024conceptgraphs, yamazaki2024open}, navigation~\cite{DBLP:journals/corr/lelan2024, Sridhar2023NoMaDGM, DBLP:conf/cvpr/Li0MIVL024, tian2024tokenize, DBLP:conf/cvpr/ShaoHWSW0024}, and robotic manipulation~\cite{DBLP:RT12023, DBLP:RT22023, DBLP:RTH2024, DBLP:RoboAgent_2024, DBLP:Octo2024, DBLP:OpenVLA2024, DBLP:RoboFlamingo2024}. To enhance grounding, systems like OpenVLA~\cite{DBLP:OpenVLA2024}, Octo~\cite{DBLP:Octo2024}, and ECoT~\cite{zawalski2024robotic} leverage powerful pretrained encoders such as DINOv2~\cite{oquab2023dinov2} and SigLIP~\cite{siglip2023} for image understanding. 

However, despite their success in reactive policy learning, most VLA models operate under an image-based Markovian assumption, treating each observation-action pair independently and have not explicitly modeled interaction history or temporal dependencies. As demonstrated in our experiments, existing VLA models perform poorly on LIBERO-Mem, where partial observability and observation aliasing challenge purely reactive, observation-driven policies. This reveals a fundamental limitation in current VLA architectures: the absence of structured memory and temporal reasoning capabilities necessary for long-horizon control. In addition to introducing LIBERO-Mem as a targeted benchmark to expose these gaps, we also propose an initial solution, grounded in object-centric modeling, that incorporates memory into the policy architecture and shows promising gains under these more realistic, memory-intensive settings.

\vspace{1em}
\noindent \textbf{Object-Centric Learning in Vision and Robotics.}
Object-centric learning has emerged as a powerful paradigm for extracting modular, interpretable representations and modeling their dynamics across space and time~\cite{GoyalLHSLBS21rim,JiangDSA23slotdiffusion}. Both supervised and unsupervised methods~\cite{savipp2022,actionslot2024, unsupervised-openvocab-slot2023iccv,uno} aim to decompose visual inputs into discrete, entity-centric representations, enabling structured reasoning, temporal modeling, and generalization across scenes. A common design across these models involves learning a fixed set of latent “slots” or components that dynamically bind to visual entities \cite{locatello2020object,mondal2024slot}, facilitating object-level understanding and forecasting. Such representations have proven effective in structured visual environments and have been extended to sequential settings where modeling object dynamics over time is crucial.

In robotic manipulation, object-centric representations help decompose complex scenes into manipulable entities, allowing for more structured policy learning and goal conditioning. Early approaches often relied on pose-based or category-specific object definitions~\cite{devin2018deep,migimatsu2020object,tyree20226}, but their dependence on supervision limits generalization to novel settings. Recent unsupervised methods aim to segment visual inputs into object-like regions~\cite{locatello2020object,heravi2022visuomotor}, offering more flexibility, but they still struggle in cluttered scenes or visually identical objects. Importantly, while these approaches facilitate object-aware perception and control, they are typically designed for short-horizon or episodic tasks, and do not incorporate mechanisms for tracking object identity and state over extended temporal sequences, a capability essential for reasoning in non-Markovian settings.

To address this gap, we take inspiration from cognitive science theories of human reasoning over discrete, persistent objects~\cite{Lam2020MentalRO}, as well as recent advances in modular, slot-based architectures~\cite{jiang2024slot, actionslot2024}. We introduce Embodied-SlotSSM, a memory-centric model that extends object-centric learning to long-horizon manipulation tasks. By explicitly modeling the temporal evolution of individual object states, our approach supports structured memory retention over time, enabling more robust reasoning and decision-making in partially observable, non-Markovian environments as exemplified in LIBERO-Mem.

\begin{table}[!h]
    \centering
     \caption{Design factors in robot manipulation benchmarks.}
    \scalebox{0.65}{ 
    \setlength{\tabcolsep}{3pt}
    \begin{tabular}{p{4.7cm} | c | c c c  c}
        \toprule
        \textbf{Benchmark feature} & \textbf{\shortstack{LIBERO\\-Mem}} & \textbf{\shortstack{Memory\\Bench}} & \textbf{\shortstack{MIKASA\\-Robo}} & \textbf{LIBERO}  & \textbf{RLBench} \\
        \midrule
        Non-Markovian observations              & \cmark & \cmark & \cmark & \xmark  & \xmark \\
        Long-horizon trajectories               & \cmark & \cmark & \xmark & \cmark 
        & \xmark \\
        Subgoal-aware evaluation     & \cmark & \xmark & \xmark & \xmark  & \xmark \\
        Object identity ambiguities        & \cmark & \xmark & \xmark & \xmark  & \xmark \\
        Object \& subgoal annotations & \cmark & \xmark & \xmark & \xmark  & \xmark \\
        Stress-testing by temporal scaling     & \cmark & \xmark & \xmark & \xmark  & \xmark \\
        \bottomrule
    \end{tabular}}
    \label{tab:benchmark_comparison}
\end{table}
\section{Non-Markovian Robot Manipulation}
\label{sec:nmmbench}
\noindent \textbf{Preliminaries.}
Instruction-conditioned robotic policies are typically trained to map a sequence of sensory inputs and a language goal into a sequence of actions. A common formulation assumes that the agent can infer the optimal next action from its current visual observation and the instruction alone. While this simplification enables efficient learning and inference, it neglects the temporal structure inherent in many real-world tasks, where past interactions and object histories are essential for disambiguating the current state. Formally, we can consider VLA learning via a demonstration dataset,
\begin{equation}
\mathcal{D} = \{ l, \mathbf{a}_{1:T}, \mathbf{v}_{1:T} \},
\end{equation}
where $l$ is a natural language instruction, $\mathbf{a}_{1:T} = \{\mathbf{a}_1, \ldots, \mathbf{a}_T\}$ is a sequence of actions, and $\mathbf{v}_{1:T} = \{\mathbf{v}_1, \ldots, \mathbf{v}_T\}$ is a sequence of visual observations. 

Many VLA models~\cite{DBLP:OpenVLA2024, zawalski2024robotic} are trained to predict the next action using only the current observation and instruction:
\begin{equation}
\hat{\mathbf{a}}_t \sim P_\theta\big(\mathbf{a}_t \mid \mathbf{v}_t, l\big),
\end{equation}
which implicitly assumes a Markovian process:
\begin{equation}
P\big(\mathbf{a}_t \mid \mathbf{v}_{1:t}, l\big) = P\big(\mathbf{a}_t \mid \mathbf{v}_t, l\big).
\end{equation}

However, in many practical scenarios, such as cooking, laboratory automation, and industrial assembly, his assumption fails. Robots frequently operate under partial observability, where identical visual inputs can correspond to different semantic states depending on prior actions. For instance, a robot may need to remember whether it has already poured liquid into a container or completed a previous subtask. These cases highlight the limitations of reactive policies and the necessity for memory-based reasoning. This violation of the Markov assumption can be formally expressed by the existence of two timesteps $t_1$ and $t_2$ such that $\mathbf{v}_{t_1} \approx \mathbf{v}_{t_2}$, but the correct actions differ due to distinct histories:
\begin{equation}
P\big(\mathbf{a}_{t_1} \mid \mathbf{v}_{1:t_1}, l\big) \neq P\big(\mathbf{a}_{t_2} \mid \mathbf{v}_{1:t_2}, l\big).
\end{equation}
To act effectively in such settings, an agent must reason over its full interaction history $(\mathbf{v}_{1:t}, \mathbf{a}_{1:t-1})$, rather than relying solely on instantaneous observations.

\section{LIBERO-Mem: non-Markovian Benchmark}

\label{sec:bench}

\renewcommand{\arraystretch}{1.25}
\begin{table}[t]
\centering
\small
\caption{LIBERO-Mem task descriptions with subgoal structure and targeted memory dimensions.}
\rowcolors{2}{gray!10}{white}
\begin{tabular}{p{3.3cm} p{2.6cm} p{1.1cm}}
\toprule
\textbf{Task Num: Description} & \textbf{Subtask Goals} & \textbf{Types} \\
\midrule
\texttt{T1:} robot to pick up the bowl and place it back on the plate & bowl lifted $\rightarrow$ \newline bowl on plate & OM \\
\texttt{T2:} robot to lift the bottle and put it down on the plate & bottle lifted $\rightarrow$ \newline bottle on plate & OM \\
\texttt{T3:} robot to lift the bowl and place it back on the plate 3 times & bowl lifted $\rightarrow$ \newline bowl on plate $\rightarrow$ \newline $\times$ 3 & OM, OS \\
\texttt{T4:} robot to pick up the bottle and put it down on the plate 3 times & bottle lifted $\rightarrow$ \newline bottle on plate $\rightarrow$ \newline $\times$ 3 & OM, OS \\
\texttt{T4:} robot to lift the bowl and place it back on the plate 5 times & bowl lifted $\rightarrow$ \newline bowl on plate $\rightarrow$ \newline $\times$ 5 & OM, OS \\
\texttt{T6:} robot to pick up the bowl and put it on the plate 7 times & bowl lifted $\rightarrow$ \newline bowl on plate $\rightarrow$ \newline $\times$ 7 & OM, OS \\
\texttt{T7:} robot to swap 2 bowls on their plates using the empty plate & bowl 1 on plate 3 $\rightarrow$ bowl 2 on plate 1 $\rightarrow$ bowl 1 on plate 2 & OM, OR \\
\texttt{T8:} robot to swap 3 bowls on their plates using the empty plate & bowl 1 on plate 4 $\rightarrow$ bowl 2 on plate 1 $\rightarrow$ bowl 3 on plate 2 $\rightarrow$ bowl 1 on plate 3 & OM, OR \\
\texttt{T9:} robot to put bowl in closest basket and move that basket to the middle & bowl 1 in basket 1 $\rightarrow$ basket 1 in center & OM, OO \\
\texttt{T10:} robot to put bowl in closest basket and move empty basket to middle & bowl 1 in basket 1 $\rightarrow$ basket 2 in center & OM, OO \\
\bottomrule
\end{tabular}
\label{tab:libero_mem_tasks}
\end{table}

\noindent \textbf{Task designs for object-centric POMDP.} LIBERO-Mem consists of 10 tasks spanning four object-centric memory dimensions: Object Motion (OM), Object Sequence (OS), Object Relations (OR), and Object Occlusion (OO). Each task presents temporal dependencies and ambiguity requiring structured memory beyond instantaneous observations. Formally, let $\mathcal{T}_i = \{o_t^i\}_{t=1}^{T}$ denote a trajectory for task $i$, where $o_t^i$ is the visual observation at time $t$. For each task $i$, there exist at least two trajectories $\mathcal{T}_i^{(1)}$, $\mathcal{T}_i^{(2)}$ such that $\exists\, t_1, t_2$ with $o_{t_1}^{(1)} = o_{t_2}^{(2)}$, yet their underlying task states or required actions differ. This guarantees that visual observations are not uniquely predictive of the correct behavior, thereby enforcing the need for structured memory across time.

\noindent \textbf{Short- and long-horizon data collection process.} Expert demonstrations are collected via smooth keyboard control with multi-key tracking. Each task contains 200--700 frames supporting both short- and long-horizon evaluation. Each task is collected to 120 trajectories, where 100 are refined as training data, and 20 are left for validation.

\noindent \textbf{Subgoal-aware evaluation.} Each task is decomposed into symbolic subgoals using Sequence ($\rightarrow$) and Or ($\vee$) operators for fine-grained evaluation. 

\noindent \textbf{Object identity ambiguities.} Visually identical bowls and plates differ only in asset ID, requiring agents to resolve object identity from temporal interaction history. See extended version for full asset details.

\noindent \textbf{Object \& subgoal annotations.} Each timestep includes object instance IDs, masks, and subgoal completion flags per object. Annotation schema is detailed in the extended version.

\noindent \textbf{Can VLAs be scaled temporally for memorization?} OpenVLA encodes entire video sequences using 256 dense tokens, while our customized object-centric VLA design operates with only 16 slot tokens, but tokens scale linearly with slot and sequence dimension, leading to intractable memory and compute costs in long-horizon settings. Thus, it motivates us to design a more scalable strategy as Embodied-SlotSSM.

\begin{figure*}[t!]
\includegraphics[width=\linewidth]{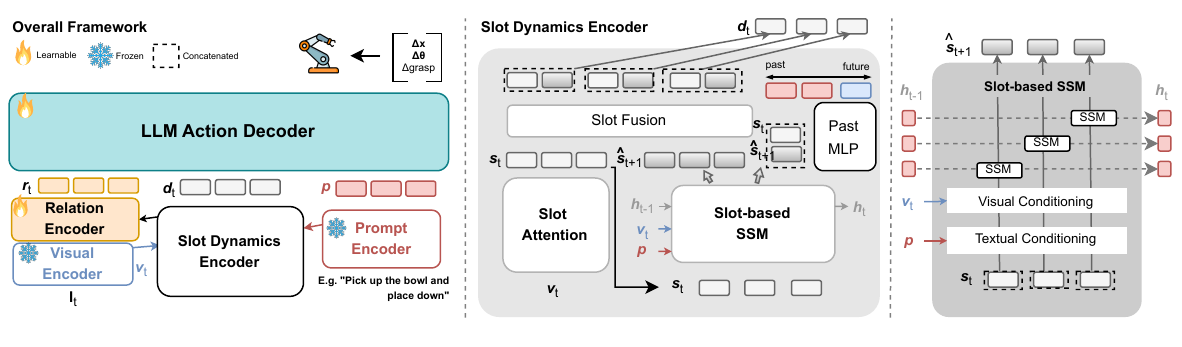}
    \caption{\textbf{Embodied-SlotSSM}: Our framework combining slot-based dynamics (Slot Attention, Slot Fusion, Slot-based SSM) with an LLM Action Decoder for object memory-aware action prediction based on textual prompts.}
\label{fig:proposed-method}
\end{figure*}


\section{Embodied-SlotSSM}
\label{sec:method}
\textbf{SlotSSM Formulation.} An SSM~\cite{mamba, mamba2} aims to approximate $\mathbf{H}_{1:t}$ as $\mathbf{h}_t$ through selective state-space modeling. In particular, the model defines a mapping from an input sequence $\mathbf{e}_{1:T} \in \mathbb{R}^{T \times D}$, encoded from $\mathbf{e}_t = v(\mathbf{o}_t)$ using a pretrained visual encoder $v(\cdot)$, to an output sequence $y_{1:T} \in \mathbb{R}^{T \times D}$ through the following input-dependent recurrence:
\begin{equation}
   \mathbf{h}_t = \overline{A}(\mathbf{e}_t) \mathbf{h}_{t-1} + \overline{B}(\mathbf{e}_t) \mathbf{e}_t, \quad y_t = C(\mathbf{e}_t) \mathbf{h}_t 
\end{equation}
where $\mathbf{h}_t \in \mathbb{R}^H$ is the hidden state summarizing the history up to time $t$, and $\overline{A}(\mathbf{e}_t) \in \mathbb{R}^{H \times H}$, $\overline{B}(\mathbf{e}_t) \in \mathbb{R}^{H \times D}$, and $ C(\mathbf{e}_t) \in \mathbb{R}^{D \times H}$ are input-conditioned matrices generated by learnable functions applied to $\mathbf{e}_t$. 
By conditioning the dynamics on the input $\mathbf{e}_t$ at every step, the model flexibly adapts its internal transitions and output mappings based on the current context.  

As the underlying process of visual representations is inherently modular, SlotSSM~\cite{jiang2024slot} is coupled with a slot encoder~\cite{locatello2020object} to decompose $\mathbf{e}_t$ into $K$ individual slot representations $\mathbf{e}_t = \left[\mathbf{s}^1_t,...,\mathbf{s}^K_t \right]$, thereby maintaining $\mathbf{h}_t$, $\mathbf{y}_t$ also modularly as $\mathbf{h}_{t} = \text{concat} \left[\mathbf{h}^1_{t},...,\mathbf{h}^K_{t}\right]$, and $\mathbf{y}_{t} = \text{concat} \left[\mathbf{y}^1_{t},...,\mathbf{y}^K_{t}\right]$. Thus, the matrices $\overline{A}_t$, $\overline{B}_t$, and $C_t$ are designed to be block-diagonal, where each block is conditioned solely on the corresponding slot input. In our work, $K = 16$ by default.
\begin{equation}
\resizebox{\linewidth}{!}{$
\overline{A}_t = \text{diag}\left( \left\{ \overline{A}(\mathbf{s}_t^k) \right\}_{k=1}^K \right), \quad
\overline{B}_t = \text{diag}\left( \left\{ \overline{B}(\mathbf{s}_t^k) \right\}_{k=1}^K \right), \quad
C_t = \text{diag}\left( \left\{ C(\mathbf{s}_t^k) \right\}_{k=1}^K \right)
$}
\end{equation}

\noindent \textbf{Embodied-SlotSSM.} Our proposed approach is a slot-based state-space model designed to enable structured, persistent memory for long-horizon visuomotor control. Thus, we make use of individual slot representations $\left[\mathbf{h}^1_t,...,\mathbf{h}^K_t\right]$ for efficient action prediction $\hat{\mathbf{a}}_t \sim P_\theta(\mathbf{a}_t | \mathbf{h}^1_t,...,\mathbf{h}^K_t, l)$ in a non-Markovian manner. To facilitate action prediction from modular representations and mitigate the memory recall challenges of SSMs~\cite{mambaempirical2024, AroraEZTA0RR24}, we propose to model both transient and persistent memory for object-centric representations that can be used for action decoding.

\subsection{Transient Memory via Temporal Localization}

To support short-term reasoning for motion encoding, we model \textit{transient memory} by temporally localizing objects using a combination of Slot Attention and State-Space Modeling. This approach enables the agent to bind scene features to discrete object-centric representations and track their temporal evolution for recent interaction history.

\subsubsection{Slot Attention for Object Localization.}
Manipulation requires disentangling and tracking discrete object entities, yet conventional visual encoders often produce entangled representations that conflate background and object-level signals. To address this, we apply Slot Attention~\cite{locatello2020object} as a tokenization function $g(\cdot)$ that transforms dense visual embeddings $\mathbf{v}_t \in \mathbb{R}^{K \times D_{\mathtt{enc}}}$ into a set of modular object-centric tokens $\mathbf{s}_t = \{\mathbf{s}_t^1, \ldots, \mathbf{s}_t^N\}$, where $\mathbf{s}_t \in \mathbb{R}^{N \times D_{\mathtt{slot}}}$. Slot Attention iteratively binds spatial feature patches to a fixed number of learnable object queries via attention and recurrent updates, yielding disentangled representations that reflect the semantics of $\mathbf{v}_t$~\cite{xu2022groupvit,DBLP:conf/iclr/JiaLH23}. The slot update at time $t$ is computed via multi-head attention followed by a recurrent refinement:
\begin{equation}
\label{eq:slot_attention}
\begin{aligned}
    &\mathbf{a}_{i,j} = \frac{1}{\sqrt{D_{\mathtt{enc}}}} \mathbf{q}_i \cdot \mathbf{k}_j^\top, \quad \Tilde{\mathbf{a}}_{i,j} = \frac{e^{\mathbf{a}_{i,j}}}{\sum_{l=1}^N e^{\mathbf{a}_{l,j}}}, \\
    &\mathbf{w}_{i,j} = \frac{\Tilde{\mathbf{a}}_{i,j}}{\sum_{l=1}^{K} \Tilde{\mathbf{a}}_{i,l}}, \quad \mathbf{u}_i = \sum_{j=1}^{K} \mathbf{w}_{i,j} \mathbf{v}_j, \\
    &\mathbf{s}_t^i = \text{GRU}(\mathbf{u}_i, \mathbf{s}_t^i),
\end{aligned}
\end{equation}

\noindent where $\mathbf{q}_i$, $\mathbf{k}_j$, and $\mathbf{v}_j$ denote the projected queries, keys, and values. Slot representations are refined by aggregating the input features $\mathbf{v}_t$ and updating via a GRU.

To enable temporal coherence in object identity, we initialize the slots at each timestep $t$ as:
\begin{equation}
\label{eq:slot_init}
    \mathbf{s}_t^{(0)} = 
    \begin{cases}
        \text{RandomInit}(), & \text{if } t = 0 \\
        \mathbf{s}_{t-1}^{(T)}, & \text{if } t > 0,
    \end{cases}
\end{equation}
\noindent where $T$ denotes the number of recurrent refinement steps per frame. At the beginning of a sequence ($t = 0$), slots are randomly initialized. For all subsequent steps ($t > 0$), the slots are initialized using the final outputs from the previous timestep. This allows the model to propagate slot identity across time and facilitates tracking of persistent objects.

\paragraph{Temporal Contrastive Loss.}

To further enforce temporal consistency, we employ a contrastive objective over a fixed temporal horizon. Given an anchor slot $\mathbf{s}_t^i$, we treat its corresponding slot in a nearby frame $\mathbf{s}_{t+\delta}^i$ (within the same sequence) as a \textit{positive} and contrast it against slots from different videos or different locations as \textit{negatives}. The contrastive loss is given by:
\begin{small}
\begin{align}
\label{eq:contrastive_loss}
\mathcal{L}_{\text{contrast}} = 
- \sum_{(i,t)} \log 
\frac{
\sum\limits_{(i',t') \in \mathcal{P}(i,t)} 
\exp\left( \frac{\text{sim}(\mathbf{s}_t^i, \mathbf{s}_{t'}^{i'})}{\tau} \right)
}{
\sum\limits_{(j,t'') \in \mathcal{P}(i,t) \cup \mathcal{N}(i,t)} 
\exp\left( \frac{\text{sim}(\mathbf{s}_t^i, \mathbf{s}_{t''}^{j})}{\tau} \right)
},
\end{align}
\end{small}

\noindent where $\text{sim}(\cdot, \cdot)$ is cosine similarity with $\tau = 1$. 

\subsubsection{Slot Dynamics Encoder with SlotSSM}

\begin{proposition}[Object history enables individuation under visual ambiguity]
Let $\mathcal{O}_t = \{ o_t^{(1)}, \ldots, o_t^{(k)} \}$ be a set of $k$ objects at time $t$, each represented by a latent $z_t^{(j)} = f_{\text{enc}}(v_t^{(j)})$ derived from the visual input $v_t^{(j)}$. Suppose that for all $i \ne j$, $v_t^{(i)} \approx v_t^{(j)}$ such that $z_t^{(i)} \approx z_t^{(j)}$ (i.e., the objects visually indistinguishable at $t$). Individuation of object identities cannot be achieved from the current frame alone. To disambiguate objects, a policy $\pi(a_t \mid h_t)$ must condition on a history $h_t$ that is object-specific $\mu_t^{(j)}$.
\end{proposition}

Thus, rather than predicting a single next-step embedding, our SlotSSM is designed to predict a window of $P = p + q$ static latents from the $p$ past to $q$ future object representations centered at the current timestep. 
This reflects the observation that meaningful object behavior often unfolds over short temporal segments rather than single-frame transitions. 
By modeling a localized temporal window, SlotSSM captures motion continuity, smooths noisy transitions, and anticipates short-term dynamics. The windowed prediction provides rich supervision during training by requiring the model to reconstruct both past and future latent states, such that SlotSSM not only learns forward dynamics (e.g., anticipating object motion) but also enforces temporal consistency via backward reconstruction, thereby improving representation stability and generalization under non-Markovian conditions. 
Formally, for each slot $j$, SlotSSM predicts a window of static representations around timestep $t$ as:
\begin{equation}
\left\{ \mathbf{z}_{t+\delta}^{(j)} \right\}_{\delta = -p}^{q} = W_{\text{pred}}\left( \hat{\mathbf{s}}_{t+1} || \mathbf{s}_t^{(j)} \right),
\end{equation}

\noindent where $\mathbf{z}_{t+\delta}^{(j)}$ is the set of predicted slots representing the $j-th$ object around the time step $t$, which is obtained from a shared decoding MLP function, called \textit{Past MLP}, with weights $W_{\text{pred}}$ applied independently to each slot. This formulation allows SlotSSM to function as a transient memory module, retaining localized temporal context per object slot to support robust tracking and downstream action decoding.

\subsection{Action Control through Slot-Conditioned Decoding}
In practice with observation chunking, $P$ is not at sequence length due to compute memory limit, so for a naïve version of Embodied-SlotSSM where (Naive E-SlotSSM), we further use an oracle text-embedding subgoal $\mathbf{g}_{t}^{(j)}$ (e.g. ``bowl 1 on plate 3'' for swapping tasks like \texttt{T7}) for each relevant object. Given a slot-conditioned decoder then predicts actions from these object-centric features, SlotSSM maintains a latent state $\mathbf{d}_{t}^{(j)}$ via a \textit{Slot Fusion} module applied to $\mathbf{s}_{t}^{(j)}$, the predicted next-slot $\hat{\mathbf{s}}_{t+1}^{(j)}$ and oracle subgoal $\mathbf{g}_{t}^{(j)}$, capturing both current dynamics and temporal context at object level. 

Then, to condition action generation on both the structured memory and the current scene, we introduce a lightweight \textit{Relation Encoder} that produces 16 relational tokens (for $K = 16$ slots). This module performs cross-attention between the slot latents ${\{\mathbf{d}_t^{(j)}}\}_{j=1}^K$ and raw visual features $\mathbf{v}_t$ to produce $L$ relation tokens ${\{\mathbf{r}_t^{(j)}}\}_{j=1}^L$, enabling context-aware reasoning over object states and interactions. The final action $\hat{\mathbf{a}}_t$ is decoded through a VLA head conditioned on: (1) the relation tokens $\{\mathbf{r}_t^{(j)}\}_{j=1}^L$, (2) the slot dynamics $\{\mathbf{d}_t^{(j)}\}_{j=1}^K$, and (3) the current task query embedding $l$:
\begin{equation}
\hat{\mathbf{a}}_t \sim P_\theta\left( \mathbf{a}_t \mid \{\mathbf{r}_t^{(j)}\}_{j=1}^L, \{\mathbf{d}_t^{(j)}\}_{j=1}^K, l \right).
\end{equation}

\noindent By grounding actions in object-centric memory and task semantics, the policy gains the ability to reason over identity, relational context, and temporal dependencies, thereby supporting robust performance in partially observable, long-horizon manipulation tasks.

\section{Experiments}
\label{sec:exp}

\subsection{Experimental Setup}

We evaluate our model on object-centric manipulation tasks using the LIBERO-Goal and the LIBERO-Mem benchmark to respectively evaluate general and object-level POMDP task performance. All experiments are conducted in simulation with fixed scene layouts and randomly initialized object poses. Each task is repeated for $N$ multiple seeds, where action accuracy is computed as the ratio of successful completions over total attempts ($\text{success rate} = \frac{\text{n$_0$ success}}{N}$) in Table~\ref{tab:perf-liberogoal}, or subgoal completion $= \frac{\text{subgoals completed}}{\text{total subgoals}}$ over $N$ seeds in Table~\ref{tab:perf-liberomem}. We evaluate comparatively against Naive E-SlotSSM using OpenVLA~\cite{DBLP:OpenVLA2024} with slot attention~\cite{locatello2020object} (as object-centric version of SlotVLA~\cite{slotvla2025}), $\pi_0$~\cite{black2024pi0visionlanguageactionflowmodel}, each with horizon $h$ denoting number of input frames, over $N = 20$.

\paragraph{General Task Performance:}

Our empirical Naive E-SlotSSM achieves the highest success rates on LIBERO-Goal, outperforming slot-based baselines. The results in Table~\ref{tab:perf-liberogoal} highlight the effectiveness of structured object-centric memory in non-Markovian manipulation tasks. While SlotVLA ($h=8$) achieves moderate performance ($75.5\%$) by extending temporal context length, it still struggles on tasks involving multi-object interactions and occlusions (e.g., \textit{middle drawer open}, \textit{top drawer open $\rightarrow$ bowl in}). In contrast, Naive E-SlotSSM consistently outperforms others ($\mathbf{83.0\%}$ avg), demonstrating robustness across both simple and temporally entangled tasks. Notably, it handles action sequences involving relational reasoning (e.g., \textit{bottle in rack}) and spatial displacements (e.g., \textit{bowl in plate}) reliably, attributed to its ability to integrate persistent slot tracking memory. These results affirm the usefulness of memory design for generalizing over object-centric tasks in general manipulation.

\begin{figure*}[t!]
\centerline{\includegraphics[width=0.85\linewidth]{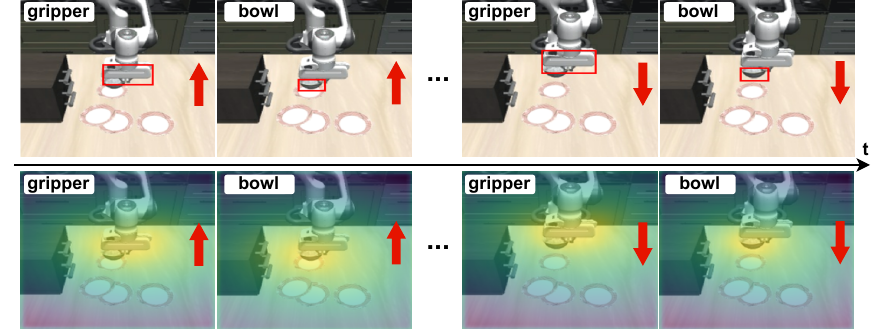}}
    \caption{Slot visualization in task \texttt{T1}: gripper and bowl slots (bbox, attention) as robot lifts and places the bowl down.} 
\label{fig:slot-vis}
\end{figure*}

\paragraph{Object POMDP Task Performance with LIBERO-Mem:}

The empirical Naive E-SlotSSM also achieves the highest subgoal success rates on LIBERO-Goal, outperforming the baselines.
Table~\ref{tab:perf-liberomem} reveals the difficulty of fine-grained subgoal tracking in non-Markovian object manipulation. Both dense-token baseline ($\pi_0$) and SlotVLA models ($h=1$, $h=8$) achieve minimal subgoal completion (avg. $5.0\%$), failing to reason over sequences such as repeated placements or multi-step swaps. In contrast, Embodied-SlotSSM yields a significantly higher average subgoal completion of $14.8\%$, with partial successes in long-horizon repetitive actions (e.g., $50\%$ on \texttt{T1} of pick and place bowl (1x), and $33.3\%$ on the repeated (3x) version. These gains suggest that slot-based temporal memory, especially when aligned with object and subgoal states, provides a strong inductive bias for step-wise reasoning under partial observability. Nonetheless, as persistent memory modeling remains modest by relying on an object-level subgoal monitor to track progress and monitor progress performance, indicating that subgoal grounding is a key open challenge in real-world non-Markovian settings.

\begin{table}[t!]
\centering
\caption{Task success rates on LIBERO-Goal.}
\centering
\resizebox{1.0\linewidth}{!}{%
\begin{tabular}{l | c c c c}
\toprule
\textbf{Task} &
\textbf{SlotVLA (h=1)} &
\textbf{SlotVLA (h=8)} &
\textbf{Naive.~SlotSSM} \\
{(token n$_0$)} & {(16)} & {(128)} & {(32)} \\
\midrule
bowl in stove & 45\% & 95\% & 100\% \\
bowl in cabinet & 75\% & 90\% & 90\% \\
plate on front-of-stove & 70\% & 80\% & 95\% \\
stove turned on & 70\% & 95\% & 100\% \\
bottle on cabinet & 20\% & 90\% & 90\% \\
bowl on plate & 0\% & 90\% & 90\% \\
mid drawer open & 5\% & 25\% & 45\% \\
cheese in bowl & 25\% & 55\% & 75\% \\
bottle on rack & 10\% & 70\% & 75\% \\
top drawer open $\rightarrow$ bowl in & 0\% & 65\% & 70\% \\
\midrule
\textbf{Average} & 32\% & 75.5\% & \textbf{83.0\%} \\
\bottomrule
\end{tabular}
\label{tab:perf-liberogoal}
}
\end{table}

\paragraph{Qualitative Analyses:}

Like SlotVLA~\cite{slotvla2025}, we visualize slot attention across time to observe whether the model consistently attends to the same object instance, especially under occlusion or motion. Visualizations (Figure~\ref{fig:slot-vis}) reveal that our model maintains consistent attention to target objects over time. This indicates the emergence of robust object permanence and tracking, which can be critical for long-horizon reasoning. More qualitative analyses of the trajectories will be shown in the extended version.

\begin{table}[t!]
\centering
\caption{Subgoal completion percentages on LIBERO-Mem.}
\centering
\resizebox{1.0\linewidth}{!}{%
\begin{tabular}{c | c c c c}
\toprule
\textbf{Task} &
$\boldsymbol{\pi}_0$ \textbf{(h=1)} &
\textbf{SlotVLA (h=1)} &
\textbf{SlotVLA (h=8)} &
\textbf{Naive.~SlotSSM} \\
{(token n$_0$)} & {(256)} &
{(16)} &
{(128)} &
{(32)} \\
\midrule
\texttt{T1} & 50.0\% & 0\% & 50.0\% & 50.0\% \\
\texttt{T2} & 0\% & 0\% & 0\% & 0\% \\
\texttt{T3} & 0\% & 0\% & 0\% & 33.3\% \\
\texttt{T4} & 0\% & 0\% & 0\% & 0\% \\
\texttt{T5} & 0\% & 0\% & 0\% & 14.3\% \\
\texttt{T6} & 0\% & 0\% & 0\% & 0\% \\
\texttt{T7} & 0\% & 0\% & 0\% & 0\% \\
\texttt{T8} & 0\% & 0\% & 0\% & 0\% \\
\texttt{T9} & 0\% & 0\% & 0\% & 30\% \\
\texttt{T10} & 0\% & 0\% & 0\% & 20\% \\
\midrule
\textbf{Average} & 5.0\% & 0\% & 5.0\% & \textbf{14.8}\% \\
\bottomrule
\end{tabular}
\label{tab:perf-liberomem}
}
\end{table}

\section{Conclusion}
\label{sec:conclusion}
The complexity of real-world environments requires agents to reason over past interactions, especially when handling multiple similar objects. Such settings demand persistent object-specific memory and introduce non-Markovian dependencies beyond reactive control.
To study this challenge, we introduced LIBERO-Mem, a benchmark for memory-intensive robotic manipulation featuring long-horizon, temporally entangled, and repetitive subgoals that test robust memory rather than short-term perception.
We further proposed Embodied-SlotSSM, a structured memory model maintaining spatio-temporally consistent slot representations that track object identity and state, thereby enabling temporally grounded, context-aware action prediction. Experiments on LIBERO-Mem show that the proposed method surpasses memory-less baselines, thereby advancing object-centric decision-making in non-Markovian environments and encouraging scalable memory-based VLAs grounded in structured representations.

\noindent \textbf{Limitations:} LIBERO-Mem is a simulated setting for future physical extension. The empirical version of Embodied-SlotSSM (Naive E-SlotSSM) currently serves as a weak baseline that leverages oracle subgoal representations rather than inferring it autonomously, leaving open the challenge of self-discovered subgoal reasoning and POMDP at object level. 

\noindent \textbf{Broader Impact:} LIBERO-Mem and Embodied-SlotSSM advance memory-aware, object-centric visuomotor systems for real-world settings, helping robots track task structure and avoid redundant actions in daily and industrial settings.

\bibliography{main}

\clearpage
\setcounter{page}{1}
\setcounter{section}{0}
\setcounter{figure}{0}
\setcounter{footnote}{0}
\setcounter{table}{0}

\definecolor{mypink}{RGB}{255, 234, 229}
\definecolor{myblue}{RGB}{220, 234, 247}
\definecolor{mygray}{RGB}{217, 217, 217}

\section{Appendix}
\label{sec:appendix}

We provide additional details on dataset usage and computational experiments to support reproducibility of our LIBERO-Mem benchmark and Embodied-SlotSSM framework. 
Readers are encouraged to check our extended version for any updates, clarifications, and additional analyses released after publication.

\subsection{Dataset Usage}

\paragraph{Motivation for Dataset Selection.}  
We evaluate our method on two benchmark families: the original LIBERO-Goal benchmark and our proposed LIBERO-Mem suite.  
LIBERO-Goal~\cite{libero2023} contains Markovian manipulation tasks and serves as a baseline for standard visuomotor performance.  
LIBERO-Mem introduces non-Markovian, memory-centric tasks that stress-test object permanence, temporal dependencies, relational reasoning, and occlusion.  
These datasets together enable evaluating how Embodied-SlotSSM handles both conventional and memory-dependent visuomotor challenges.

\begin{figure}[h!]
\centerline{\includegraphics[width=1.0\linewidth]{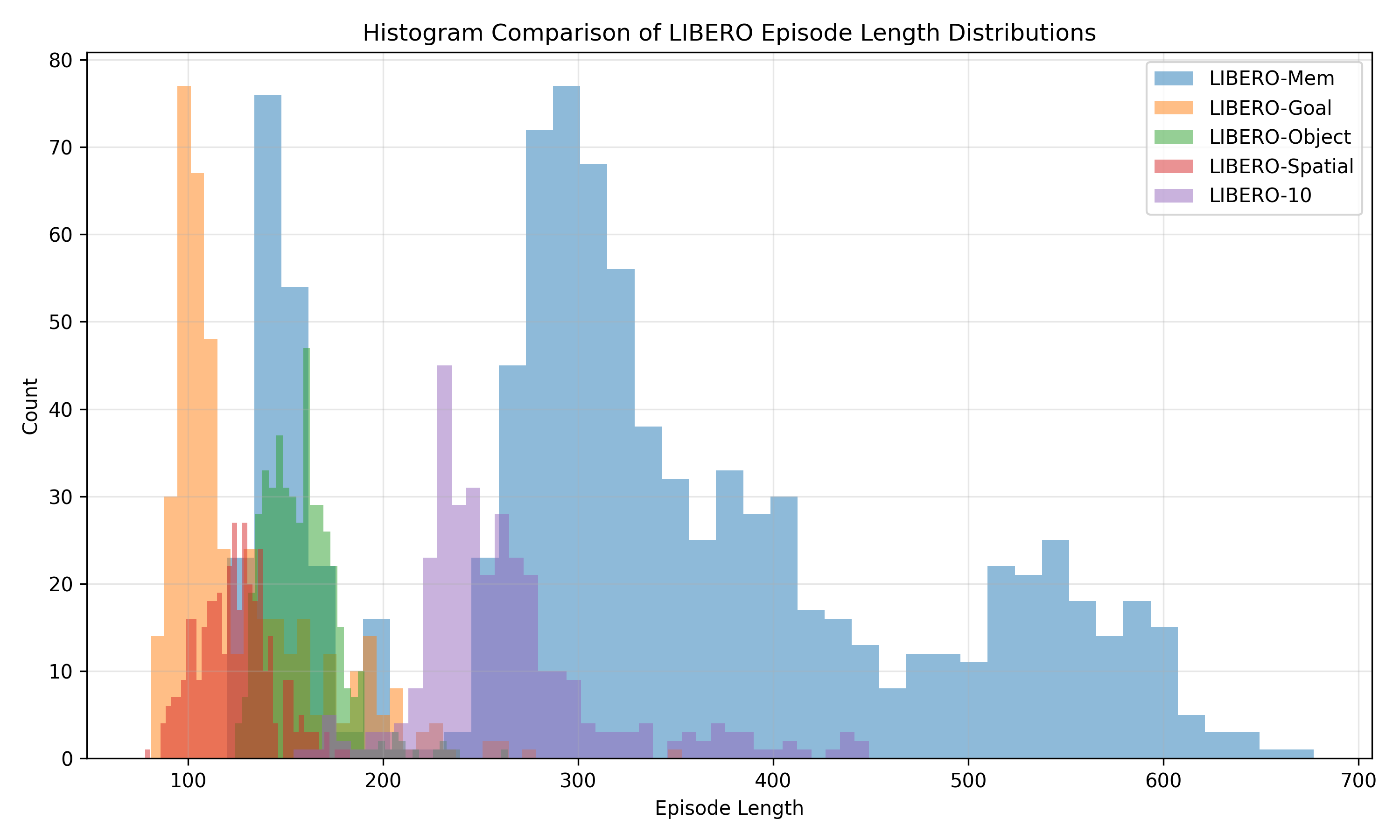}}
    \caption{The histogram of frame count across different subsets of LIBERO and our proposed LIBERO-Mem.} 
\label{fig:appen-hist-overlaid}
\end{figure}

\begin{figure}[t!]
\centerline{\includegraphics[width=1.0\linewidth]{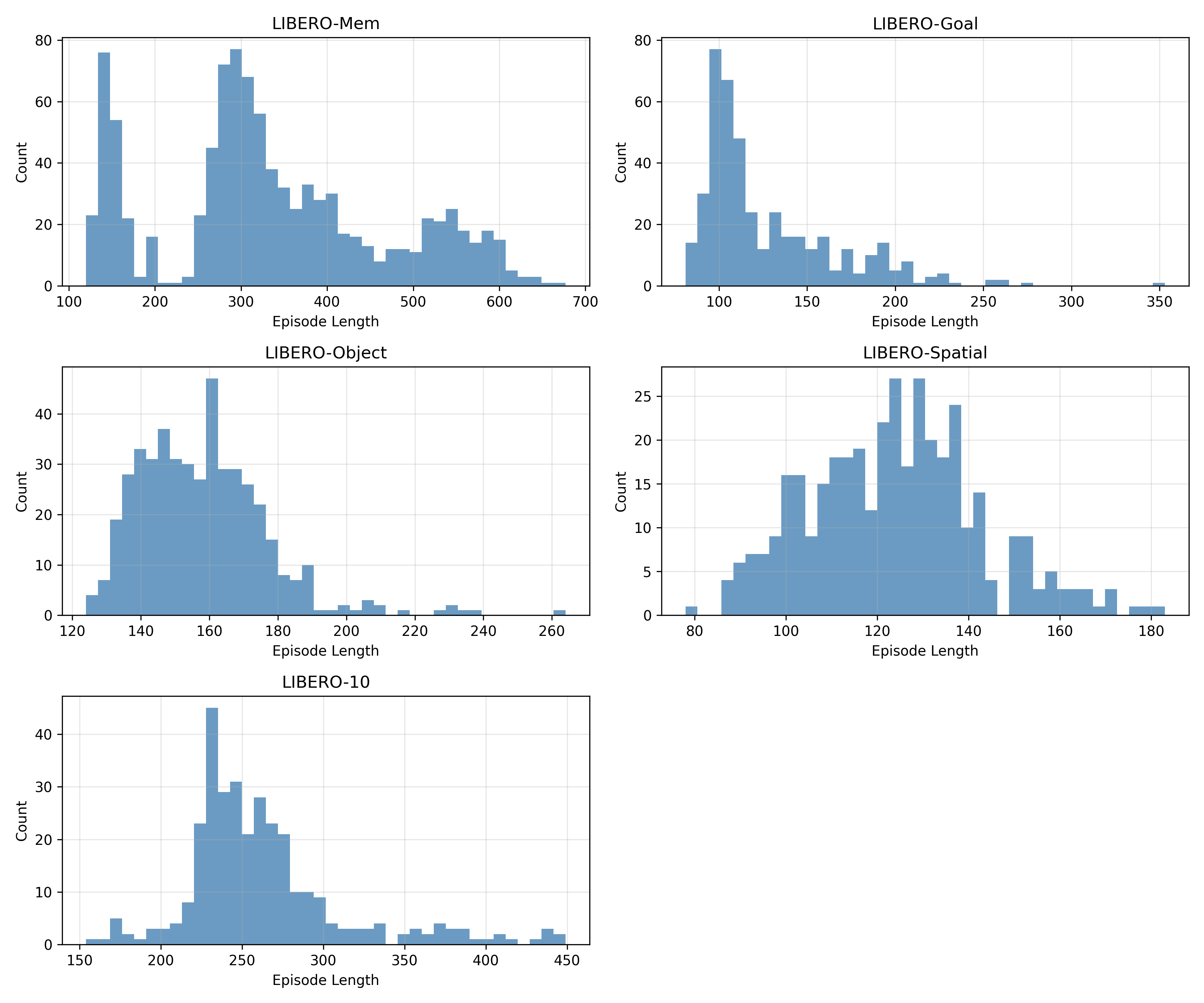}}
    \caption{The detailed histograms of frame count across different subsets of LIBERO and our proposed LIBERO-Mem.} 
\label{fig:appen-hist-subplots}
\end{figure}

\begin{figure*}[h!]
\centerline{\includegraphics[width=1.0\linewidth]{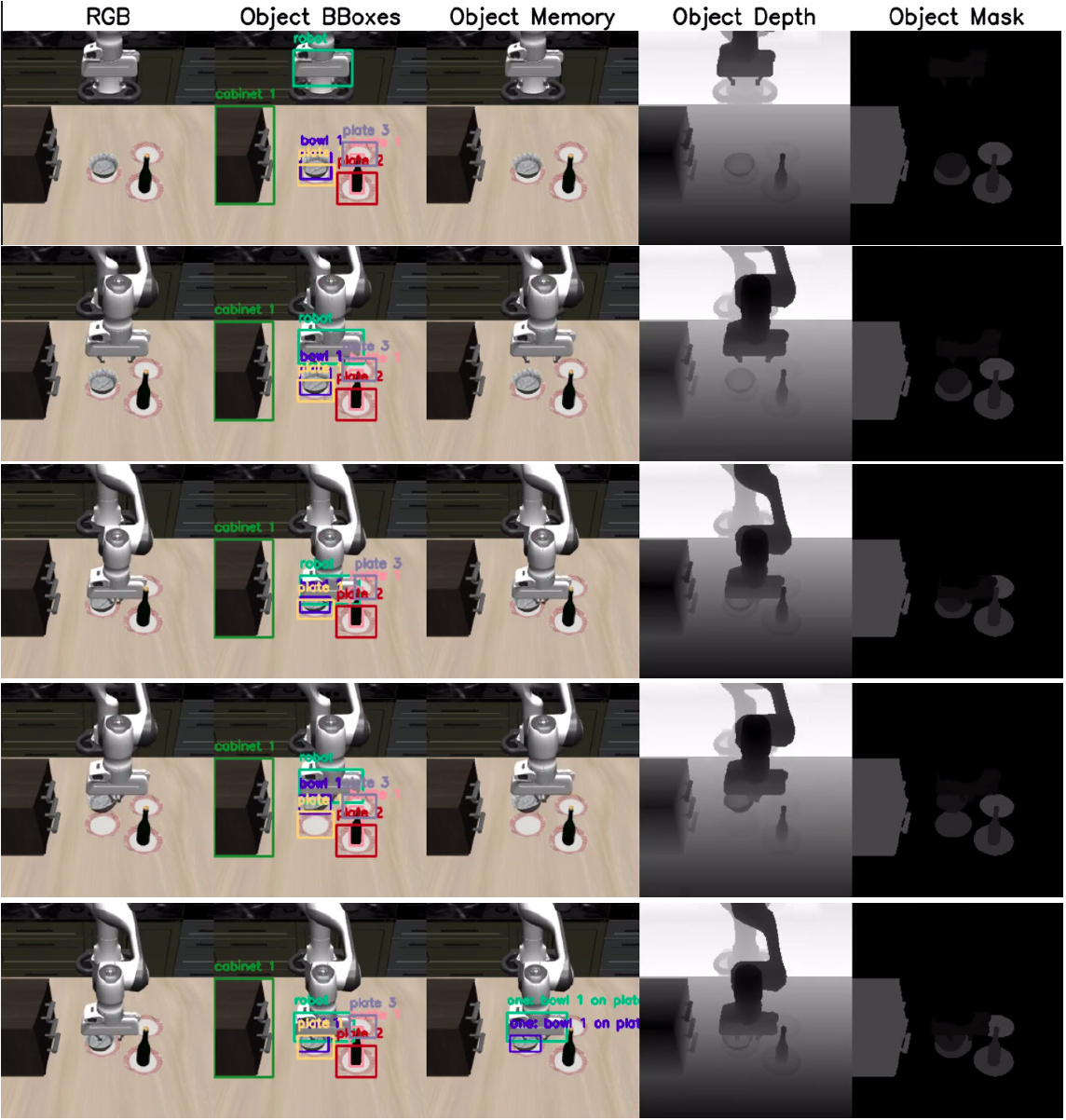}}
    \caption{Example of OM setting with \texttt{T1}.}
\label{fig:om}
\end{figure*}

\begin{figure*}[h!]
\centerline{\includegraphics[width=1.0\linewidth]{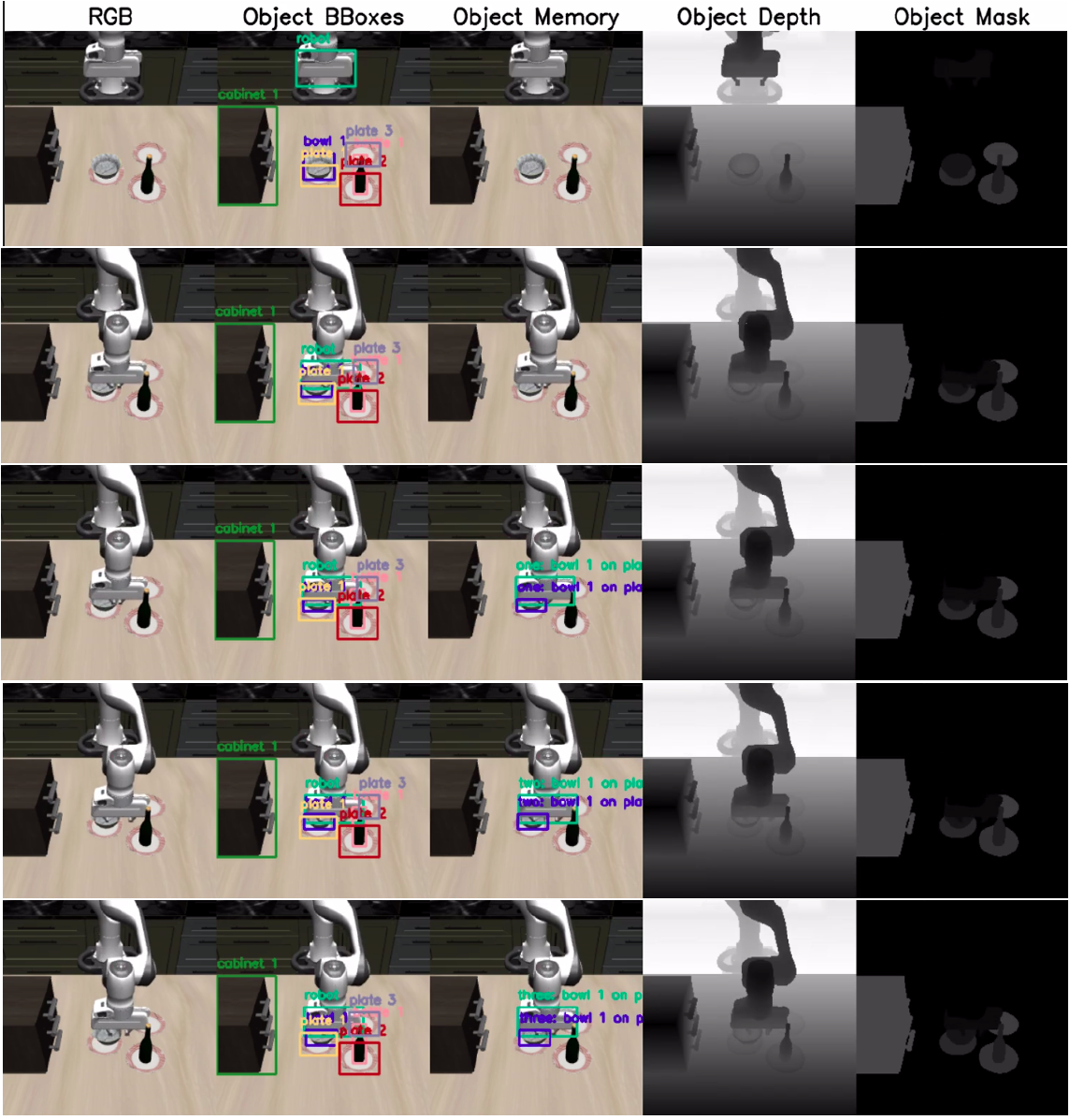}}
    \caption{Example of OS setting with \texttt{T3}.}
\label{fig:os}
\end{figure*}

\begin{figure*}[h!]
\centerline{\includegraphics[width=1.0\linewidth]{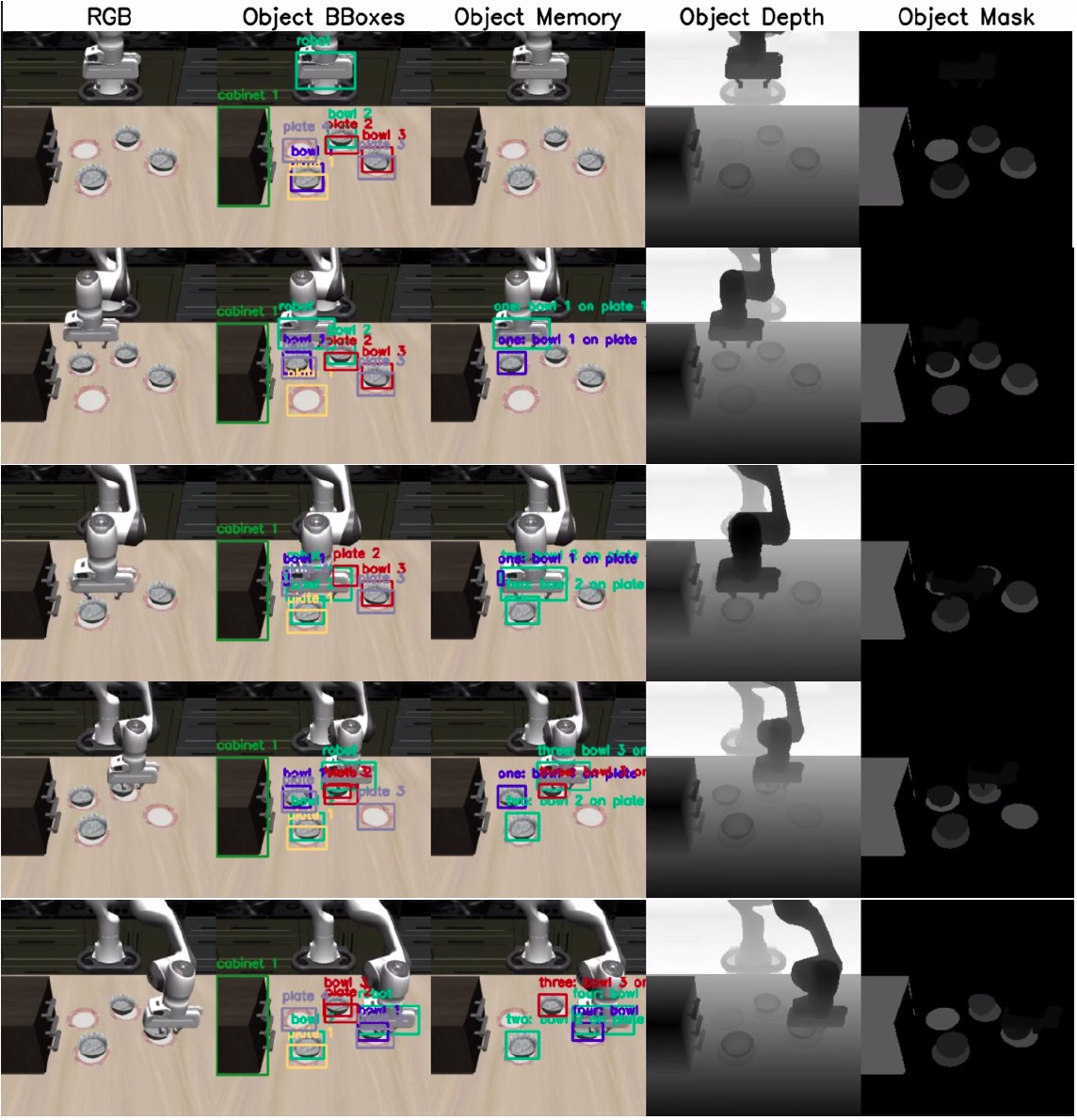}}
    \caption{Example of OR setting with \texttt{T8}.}
\label{fig:or}
\end{figure*}

\begin{figure*}[h!]
\centerline{\includegraphics[width=1.0\linewidth]{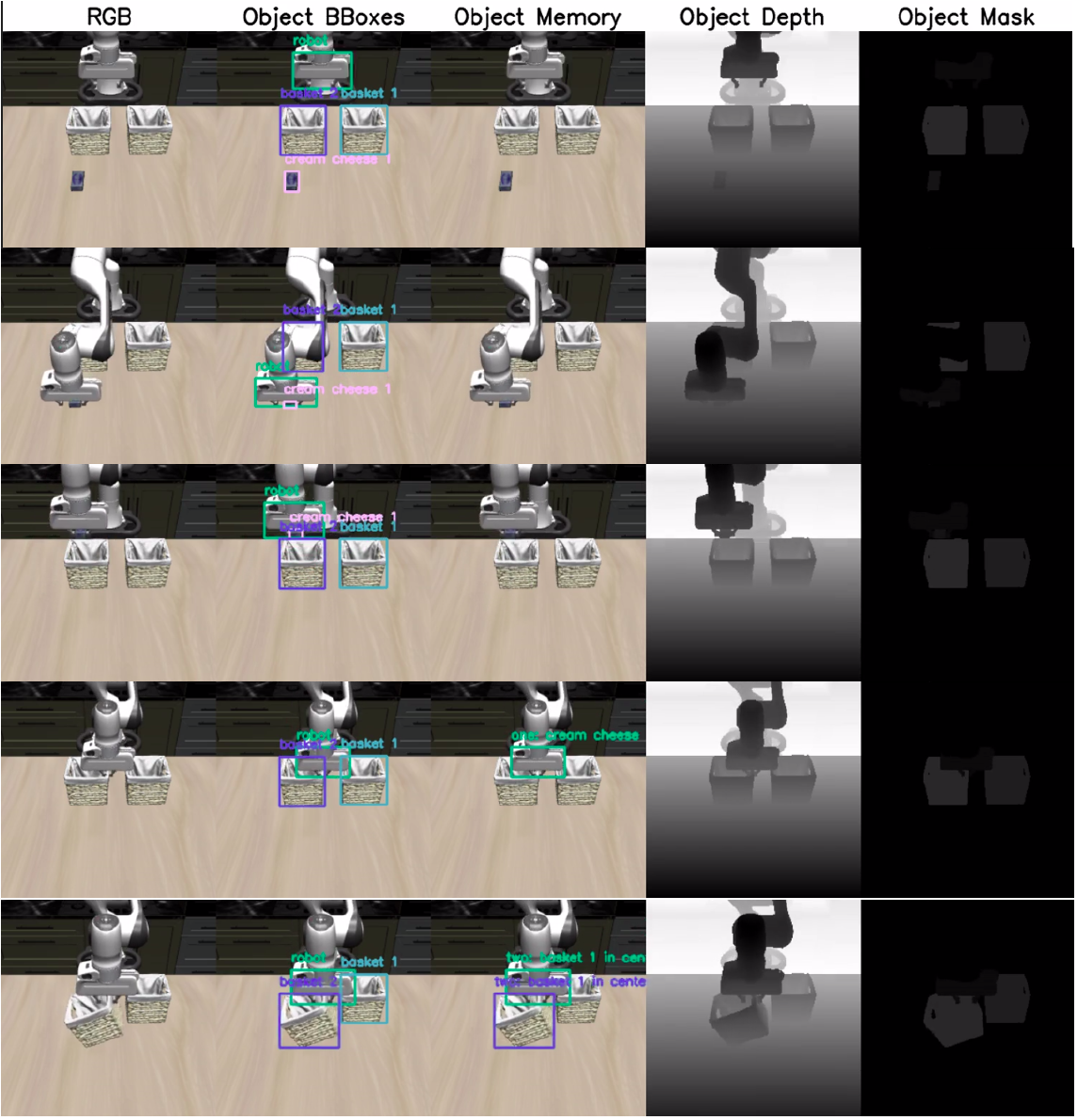}}
    \caption{Example of OO setting with \texttt{T9}.}
\label{fig:oo}
\end{figure*}

\paragraph{Novel Dataset Components.}  
LIBERO-Mem will be released publicly at a public link~\footnote{\url{https://libero-mem.github.io}} and includes ten newly constructed tasks across four memory dimensions (Object Motion, Object Sequence, Object Relations, and Object Occlusion).  
Each task is provided with metadata describing temporal dependencies, subgoal structures, and expected memory behaviors. We provide analyses of the dataset in Fig.~\ref{fig:appen-hist-overlaid} and Fig.~\ref{fig:appen-hist-subplots}. Furthermore, we show examples of OM, OS, OR, OO settings in Fig.~\ref{fig:om}, Fig.~\ref{fig:os}, Fig.~\ref{fig:or}, Fig.~\ref{fig:oo} with RGB image longside object boxes, object memory, object depth and object mask.

\paragraph{Public Release of Novel Datasets.}  
We release LIBERO-Mem with: RGB frames, object masks, subgoal labels, trajectory metadata, and task specifications.  The dataset will be accessible under a research-friendly license.~\footnote{https://huggingface.co/datasets/libero-mem/LIBERO-Mem}

\paragraph{Existing Datasets and Citations.}  
All datasets drawn from existing literature, particularly LIBERO-Goal~\cite{libero2023}, are properly cited. We follow the data setting used in OpenVLA~\footnote{https://github.com/moojink/rlds\_dataset\_builder} in data building.

\paragraph{Public Availability.}  
All datasets used in this study are publicly available or will be made public at the links provided. No proprietary or restricted-access data are used.

\paragraph{Non-Public Data.}  
There are no non-public datasets involved in the experiments presented in this paper.

\subsection{Computational Experiments}

In this section, we discuss the implementation of Naive E-SlotSSM as a representation for Embodied-SlotSSM,

\paragraph{Data Pre-processing Code.}  
We release the full pre-processing pipeline, including scripts for extracting RGB observations, aligning instance masks, generating slot tokens, and constructing subgoal labels.  
All code paths needed for reproducing LIBERO-Mem training inputs are documented in the codebase.

\begin{figure*}[h!]
\centerline{\includegraphics[width=1.0\linewidth]{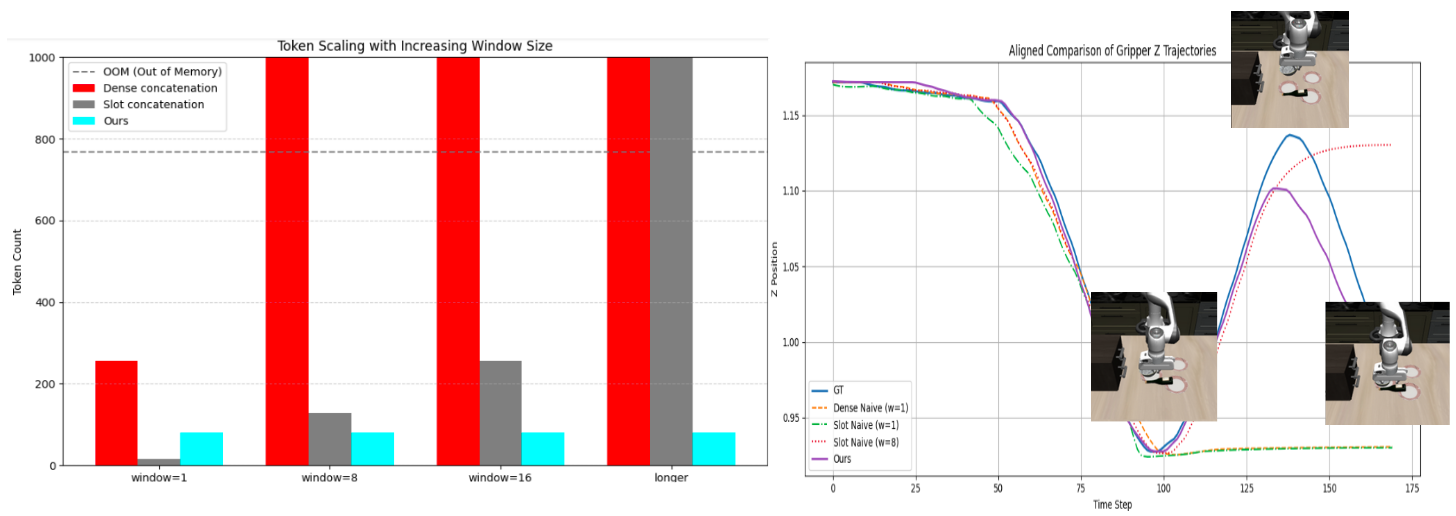}}
    \caption{Token scaling challenges under different temporal window sizes vs visual token count (left) and naive performances qualitatively shown with our simple OM task of pick and place down (right). Slot refers to use of SlotVLA~\cite{slotvla2025}, and Dense refers to use of OpenVLA~\cite{DBLP:OpenVLA2024}. We found that concatenation-based methods struggles to pick up and place down, either stuck when it needs to go up or stuck when it needs to go down, probably during training, the model is confused by same visual states but different directions. Meanwhile, our method could pick up the bowl and plate on the plate.} 
\label{fig:token-scaling}
\end{figure*}

\paragraph{Experimental Source Code.}  
The Naive Embodied-SlotSSM implementation are to be released publicly\footnote{\url{https://github.com/libero-mem/naive-e-slotssm}} and the evaluation toolkit for LIBERO-Mem are also to be released publicly\footnote{\url{https://github.com/libero-mem/libero-mem}}. Here, we also show that Naive Embodied-SlotSSM can systematically addresses the token scaling problems when memory grounding is important, as shown in Fig.~\ref{fig:token-scaling}. The evaluation algorithm is shown in Algorithm~\ref{alg:goal-eval}, and the difference between our evaluation strategy versus an existing one is shown in Table~\ref{tab:libero-compare}. In order to perform evaluation, our subgoals are exemplified as follows,

\begin{myBox}[{OM Task: pick up bowl and place it back on the plate 1 time}, size=small]
\footnotesize
(:goal\\
\quad(Sequence\\
\quad\quad(And (On akita\_black\_bowl\_1 plate\_1))\\
\quad))
\end{myBox}

\begin{myBox}[{OS Task: pick up bowl and place it back on the plate 3 times}, size=small]
\footnotesize
(:goal\\
\quad(Sequence\\
\quad\quad(And (On akita\_black\_bowl\_1 plate\_1))\\
\quad\quad(And (On akita\_black\_bowl\_1 plate\_1))\\
\quad\quad(And (On akita\_black\_bowl\_1 plate\_1))\\
\quad))
\end{myBox}

\begin{myBox}[{OR Task: swap the 2 bowls on their plates using the empty plate}, size=small]
\footnotesize
(:goal\\
\quad(Or\\
\quad\quad(Sequence\\
\quad\quad\quad(And (On akita\_black\_bowl\_1 plate\_3))\\
\quad\quad\quad(And (On akita\_black\_bowl\_2 plate\_1))\\
\quad\quad\quad(And (On akita\_black\_bowl\_1 plate\_2))\\
\quad\quad)\\
\quad\quad(Sequence\\
\quad\quad\quad(And (On akita\_black\_bowl\_2 plate\_3))\\
\quad\quad\quad(And (On akita\_black\_bowl\_1 plate\_2))\\
\quad\quad\quad(And (On akita\_black\_bowl\_2 plate\_1))\\
\quad\quad)))
\end{myBox}

\begin{myBox}[{OO Task: put the cream cheese in the nearest basket and place that basket in the center}, size=small]
\footnotesize
(:goal\\
\quad(Or\\
\quad\quad(Sequence\\
\quad\quad\quad(And (In cream\_cheese\_1 basket\_1\_contain\_region))\\
\quad\quad\quad(And (On basket\_1 kitchen\_table\_the\_center))\\
\quad\quad)\\
\quad\quad(Sequence\\
\quad\quad\quad(And (In cream\_cheese\_1 basket\_2\_contain\_region))\\
\quad\quad\quad(And (On basket\_2 kitchen\_table\_the\_center))\\
\quad\quad)))
\end{myBox}

\begin{myBox}[{OO Task: put the cream cheese in the nearest basket and place the empty basket in the center}, size=small]
\footnotesize
(:goal\\
\quad(Or\\
\quad\quad(Sequence\\
\quad\quad\quad(And (In cream\_cheese\_1 basket\_1\_contain\_region))\\
\quad\quad\quad(And (On basket\_2 kitchen\_table\_the\_center))\\
\quad\quad)\\
\quad\quad(Sequence\\
\quad\quad\quad(And (In cream\_cheese\_1 basket\_2\_contain\_region))\\
\quad\quad\quad(And (On basket\_1 kitchen\_table\_the\_center))\\
\quad\quad)))
\end{myBox}

\noindent If after the last subgoal the robot again changes the goal state, then it means the task has failed by over-repetition, scoring.

\begin{table*}[t]
\centering
\caption{Comparison between existing LIBERO goal specification and our compositional extension.}
\begin{tabular}{lll}
\toprule
\textbf{Aspect} & \textbf{Existing LIBERO} & \textbf{Ours} \\
\midrule
\textbf{Goal Type} & 
Flat conjunction (\texttt{And}) &
Composed logic (\texttt{Sequence}, \texttt{Or}, nested \texttt{And}) \\
Place bowl on plate & & \\
\midrule

\textbf{Difficulty Level} &
Low-Med: one-/two-step goals &
Low-High: multi-step, multi-object goals \\
Rearranging objects across locations \\
\midrule

\textbf{Temporal Reasoning} &
\xmark Not supported &
\cmark Sequenced steps via \texttt{Sequence} \\
Place object on 3 plates in order & & \\
\midrule

\textbf{Alternative Paths} &
\xmark One fixed goal &
\cmark Multiple valid paths via \texttt{Or} \\
Two valid sequences for table setup \\
\midrule

\textbf{Tiered Evaluation} &
\xmark Single outcome only &
\cmark Partial success with subgoal-aware evaluation \\
Ordered subgoal execution \\
\bottomrule
\end{tabular}
\label{tab:libero-compare}
\end{table*}

\begin{algorithm*}[t]
\caption{High-level hierarchical goal evaluation}
\label{alg:goal-eval}
\begin{algorithmic}[1]

\STATE \texttt{def check\_success(goal\_state, X):}
\STATE \quad \texttt{\# X: dict of object states}
\STATE \quad \texttt{if is\_structured(goal\_state):}
\STATE \quad\quad \texttt{return eval\_goal(goal\_state, X, satisfied=[])}
\STATE \quad \texttt{else:}
\STATE \quad\quad \texttt{return eval\_conjunction(goal\_state, X)}
\STATE

\STATE \texttt{def eval\_goal(G, X, satisfied):}
\STATE \quad \texttt{op = G[0]}
\STATE \quad \texttt{if op == "and":}
\STATE \quad\quad \texttt{return eval\_conjunction(G[1:], X)}
\STATE
\STATE \quad \texttt{elif op == "sequence":}
\STATE \quad\quad \texttt{k = len(satisfied)}
\STATE \quad\quad \texttt{if k >= len(G) - 1:}
\STATE \quad\quad\quad \texttt{return True}
\STATE \quad\quad \texttt{sub = G[k+1]  \# next subgoal}
\STATE \quad\quad \texttt{if eval\_goal(sub, X, satisfied):}
\STATE \quad\quad\quad \texttt{satisfied.append(sub)}
\STATE \quad\quad \texttt{return (len(satisfied) >= len(G) - 1)}
\STATE
\STATE \quad \texttt{elif op == "or":}
\STATE \quad\quad \texttt{for subseq in G[1:]:}
\STATE \quad\quad\quad \texttt{if not is\_prefix\_consistent(satisfied, subseq[1:]):}
\STATE \quad\quad\quad\quad \texttt{continue}
\STATE \quad\quad\quad \texttt{if eval\_goal(subseq, X, satisfied.copy()):}
\STATE \quad\quad\quad\quad \texttt{return True}
\STATE \quad\quad \texttt{return False}
\STATE
\STATE \quad \texttt{else:}
\STATE \quad\quad \texttt{raise ValueError("unknown operator")}
\STATE

\STATE \texttt{def eval\_conjunction(states, X):}
\STATE \quad \texttt{return all(eval\_predicate(s, X) for s in states)}
\STATE

\STATE \texttt{def eval\_predicate(state, X):}
\STATE \quad \texttt{if len(state) == 3:}
\STATE \quad\quad \texttt{p, o1, o2 = state}
\STATE \quad\quad \texttt{return eval\_pred\_fn(p, X[o1], X[o2])}
\STATE \quad \texttt{else:}
\STATE \quad\quad \texttt{p, o = state}
\STATE \quad\quad \texttt{return eval\_pred\_fn(p, X[o])}

\end{algorithmic}
\end{algorithm*}

\paragraph{Public Release of Code.}  
Upon publication, the entire codebase (data loaders, training pipelines, baseline implementations, and evaluation utilities) will remain publicly available under a research-permissive license.

\paragraph{Implementation Comments and Documentation.}  
Newly introduced components (Slot Dynamics Encoder, windowed SlotSSM predictor, slot fusion module, relation encoder, and VLA decoding head) include comments referencing their mathematical formulations. These comments are designed to support line-by-line tracing of each algorithmic step.

\paragraph{Randomness and Seed Control.}  
For experiments with stochastic behavior, we set global seeds for Python, NumPy, and PyTorch.  
Unless otherwise stated, each result averages over $N=20$ runs with distinct seeds.

\paragraph{Computing Infrastructure.}  
The models are finetuned on 2x NVIDIA A100 GPUs. To support reproducibility, the complete list of packages are included in our \texttt{requirements.txt} file included in the code repository. Readers may refer to the extended version for any additional environment details released after publication.

\paragraph{Evaluation Metrics.}  
We report:  
(1) \emph{Task Success Rate}: full completion of the LIBERO instruction.  
(2) \emph{Subgoal Completion Rate}: symbolic subgoal satisfaction using Sequence ($\rightarrow$) and Or ($\lor$) operators for LIBERO-Mem tasks. Further details of how these subgoals are set up are also provided below.
These metrics capture both overall task performance and fine-grained memory-sensitive progress.

\paragraph{Number of Runs.}  
Each experimental task configuration is evaluated over $20$ independent rollouts with different seeds.

\paragraph{Variation and Distributional Measures.}  
We report means across seeds per task or average subgoal completion in the main text. In the extended version, we aim to additionally provide standard deviations and confidence intervals to characterize performance stability.


\paragraph{Final Hyperparameters.}  
This accepted version does \emph{not} include an extended hyperparameter search.  Instead, we adopt default hyperparameters from the originally cited papers. Specifically,  
\begin{itemize}
    \item We use {$K=16$ slots} to reduce the chance of missing small or background objects.  
    \item The temporal prediction horizon is fixed to {$P=32$} frames, matching the observation chunk size during training.  
    \item All other hyperparameters follow the defaults from prior work unless explicitly noted.  
\end{itemize}

\paragraph{Hyperparameter Search Protocol.}  
Sweep-based hyperparameter search was limited as all hyperparameters are fixed from defaults or prior literature. Evaluation is conducted on a representative subset of LIBERO-Mem covering all four memory dimensions.

\end{document}